\documentclass[11pt]{article}

\usepackage[preprint]{acl}
\usepackage{tcolorbox}
\usepackage{times}
\usepackage{latexsym}
\usepackage{utfsym}
\usepackage{soul}
\usepackage[T1]{fontenc}

\usepackage[utf8]{inputenc}

\usepackage{microtype}

\usepackage{graphicx}
\usepackage{array}

\usepackage{algorithm}
\usepackage{algorithmic}
\usepackage{enumitem}
\usepackage{adjustbox}
\usepackage{tabularx}
\usepackage{inconsolata}
\usepackage{diagbox}
\usepackage{amsmath}
\usepackage{amssymb}
\usepackage{amsthm}
\usepackage{cleveref}
\usepackage{stackengine}
\usepackage{comment}
\usepackage{mathtools}
\usepackage{xcolor}
\usepackage{longtable}
\usepackage{multirow}
\usepackage{graphicx}
\usepackage{subcaption}
\usepackage{colortbl}
\usepackage{booktabs}

\theoremstyle{plain}

\usepackage{xcolor}
\usepackage{cleveref}
\usepackage{enumitem}

\definecolor{c_example}{HTML}{B2EBF2} 
\definecolor{c_question}{HTML}{F8BBD0} 
\definecolor{c_both}{HTML}{FFF9C4}    
\DeclareRobustCommand{\hlex}[1]{{\sethlcolor{c_example}\hl{#1}}}
\DeclareRobustCommand{\hlqu}[1]{{\sethlcolor{c_question}\hl{#1}}}
\DeclareRobustCommand{\hlbo}[1]{{\sethlcolor{c_both}\hl{#1}}}

\title{SAD: A Large-Scale Strategic Argumentative Dialogue Dataset}

\author{Yongkang Liu$^{1}$ Jiayang Yu$^{1}$ Mingyang Wang$^{2,3}$ Yiqun Zhang$^{1}$ Ercong Nie$^{2,3}$ \\
  \textbf{Shi Feng$^{1}$\thanks{Corresponding Author} Daling Wang$^{1}$ Kaisong Song$^{4}$ and Hinrich Schütze$^{2,3}$} \\
        $^1$Northeastern University, China;
        $^2$CIS, LMU Munich, Germany \\
        $^3$Munich Center for Machine Learning (MCML), Germany \\
        $^4$Alibaba Group, Hangzhou, China\\
        \texttt{misonsky@163.com,20221251@stu.neu.edu.cn,mingyang.wang2@de.bosch.com} \\
        \texttt{yiqunzhang@stumail.neu.edu.cn,nie@cis.lmu.de}
}

\begin{document}
\maketitle
\begin{abstract}
Argumentation generation has attracted substantial research interest due to its central role in human reasoning and decision-making. However, most existing argumentative corpora focus on non-interactive, single-turn settings, either generating arguments from a given topic or refuting an existing argument. In practice,  however, argumentation is often realized as multi-turn dialogue, where speakers defend their stances and employ diverse argumentative strategies to strengthen persuasiveness. To support deeper modeling of argumentation dialogue, we present the first large-scale \textbf{S}trategic \textbf{A}rgumentative \textbf{D}ialogue dataset, SAD, consisting of 392,822 examples. Grounded in argumentation theories, we annotate each utterance with five strategy types, allowing multiple strategies per utterance. Unlike prior datasets, SAD requires models to generate contextually appropriate arguments conditioned on the dialogue history, a specified stance on the topic, and targeted argumentation strategies. We further benchmark a range of pretrained generative models on SAD and present in-depth analysis of strategy usage patterns in argumentation.
\end{abstract}
\begin{figure}[!t]
\centering
\includegraphics[width=\columnwidth]{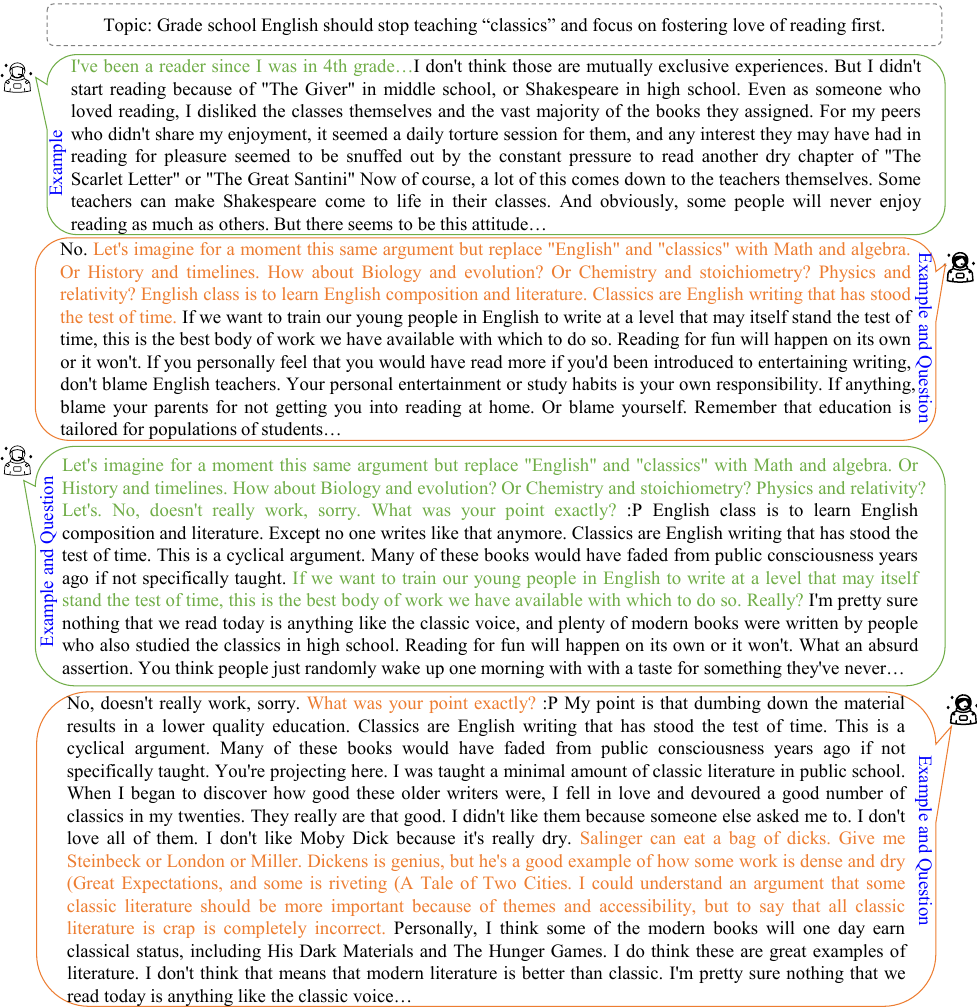}
\caption{An example of SAD. The speakers on the left are opponents of the topic, and the speakers on the right are supporters of the topic. The argumentative strategies used are highlighted in \textcolor{blue}{blue}. Strategic sentences are emphasized using other colors. }
\label{intro:example}
\end{figure}
\section{Introduction}

As fundamental capabilities of human intelligence, argumentation and debating play an important 
role in everyday activities that involve reasoning, decision making and persuasion~\citep{bar2021advances}. 
This capability is increasingly important in the era of large language models (LLMs): beyond producing fluent text, LLMs are expected to justify claims, weigh competing evidence, respond to challenges, and remain consistent under scrutiny, abilities that are naturally expressed in argumentative dialogue. 
Argumentation can be viewed as a structured exchange between supporters and opponents on a topic, typically unfolding over multiple turns.
The key to argumentation is the ability to effectively present arguments and contest an argument 
by presenting a counter-argument. The purpose of argumentation is to recognize differences and make 
the best judgment or decision through information exchange. 

Therefore, argument generation has stimulated widespread interest in the research community~\cite{durmus2019corpus,gurevychdebating,hua2019argument,ruggeri2023dataset}, with online argumentation being a major focus~\cite{tan2016winning,durmus2018exploring}. 
Two widely studied tasks are \emph{counter-argument generation} -- producing a rebuttal to a given claim, optionally conditioned on a topic~\cite{hua2019argument,alshomary2021argument,wu2022incorporating} -- and \emph{topic-conditioned argument generation} which produces arguments relevant to a given topic~\cite{sato2015end,schiller2021aspect,alshomary2023conclusion,goloviznina2023aspect}. 
However, most existing work remains \textbf{single-turn}, leaving a substantial gap to real-world argumentation, which unfolds through interactive, multi-turn exchanges.
Although the argument corpora of scientific papers include multi-turn Q\&A, each question-answer 
pair remains independent and non-interactive, essentially functioning as single-argument arguments~\citep{wu2022incorporating,ruggeri2023dataset}.
Recent attempts employ multi-agent systems to synthesize multi-turn argumentation dialogues~\cite{Li2024CanLS} , but such data does not reflect real-world interactions and is small in scale. In contrast, 
we believe that real argumentation relies heavily on strategic behavior across turns to enhance persuasiveness.

As shown in Figure~\ref{intro:example}, we present a real-world multi-turn argumentative dialogue on the topic \textit{``I believe grade school English should stop teaching `classics' and focus on fostering love of reading first.''}. The opponent believes that \textit{``teaching classics''} and \textit{``fostering love of reading''} are not contradictory by giving examples. 
Then the supporter states views to rebut by giving counter-examples and posing rhetorical question. 
Through this multi-turn exchange, both sides iteratively respond to each other’s claims and strengthen their positions. 
This example highlights three key characteristics of real argumentation: (1) it is typically an interactive dialogue between supporters and opponents with opposing stances on a topic; (2) the dialogue history provides essential context and evidence for generating coherent responses; and (3) speakers employ diverse argumentative strategies across turns to enhance persuasiveness. However, the lack of comprehensive corpora that reflect these real-world properties remains a major bottleneck for advancing argument generation.

To fill this gap, we annotate a large-scale \textbf{S}trategy \textbf{A}rgumentation \textbf{D}ialogue dataset (SAD) based on the crawled \textbf{ChangeMyView} corpus. We construct dialogue structure by tracing the interactive reply chain, yielding 722,812 utterances grouped into 392,822 dialogue examples across 20,619 topics. 
Guided by argumentation theories~\cite{freeley2013argumentation,weston2018rulebook}, linguist experts define five strategy types (Table~\ref{tab:strategy}): \textbf{Question}, \textbf{Causality}, \textbf{Example}, \textbf{Analogy}, and \textbf{Statement}. 
Each utterance is annotated with its stance (support vs.\ opposition) and may be assigned multiple strategy labels.
Based on SAD, we formulate a strategy-conditioned generation task, \textbf{P(Argument | History, Stance, Topic, Strategy)}, to better reflect real-world argumentative dialogue.
To the best of our knowledge, SAD is the first large-scale, strategy-based dataset for multi-turn argumentation. To automatically assess persuasiveness, we further train a persuasiveness evaluator using the number of likes on an argument as supervision, and show that its scores align well with human judgements.
Experimental results demonstrate that utilizing strategies improves generation quality in fluency, coherence, topicality, and persuasiveness. Finally, we present an in-depth analysis of strategy usage patterns and evaluator behavior, offering insights for building more effective argumentation dialogue systems.

\begin{table*}[!t]
\small
\centering
\resizebox{\textwidth}{!}{
\begin{tabular}{rcccccc}
\toprule
dataset & Task & Form & Number &Real World &Strategy & Multi-turn  \\
\midrule
Gigaword~\cite{sato2015end} & argument generation & R(Y|T) & 50 & \usym{1F5F8} &\usym{1F5F4} & \usym{1F5F4} \\
Iargument~\cite{wang2016neural} & argument generation & P(Y|A,T) & 18,363&\usym{1F5F8}& \usym{1F5F4} & \usym{1F5F4}\\
MedicalRecords~\cite{green2017argumentation} &argument generation &Scheme-based & -& \usym{1F5F8}& \usym{1F5F4} & \usym{1F5F4} \\
Arg-Microtexts~\cite{wachsmuth2018argumentation} & argument synthesis & -& 576&\usym{1F5F4}&\usym{1F5F8} &\usym{1F5F4}\\
CMV~\cite{hua2018neural} & argument generation & P(Y|T, KB) & 224,553&\usym{1F5F8}& \usym{1F5F4} & \usym{1F5F4} \\
Rank-30k~\cite{gretz2020workweek} & argument generation & P(Y|A,T) & 10,669&\usym{1F5F8}& \usym{1F5F4} & \usym{1F5F4} \\
CE2.3k~\cite{gretz2020workweek} & argument generation & P(Y|A,T) & 1,489&\usym{1F5F8}& \usym{1F5F4} & \usym{1F5F4} \\
LN55k~\cite{gretz2020workweek} & argument generation & P(Y|A,T) & 30,000&\usym{1F5F4}& \usym{1F5F4} & \usym{1F5F4} \\
Kialo~\cite{al2021employing} & argument generation & P(Y|T, KB) & 82,728&\usym{1F5F8}& \usym{1F5F4} & \usym{1F5F4} \\
Args.me~\cite{ajjour:2019a} & argument generation & P(Y|T, KB) & 30,748&\usym{1F5F8}& \usym{1F5F4} & \usym{1F5F4} \\
CMV~\cite{alshomary2021argument} & argument generation & P(Y|T,A) & 111,900&\usym{1F5F4}& \usym{1F5F4} & \usym{1F5F4} \\
PRRCA~\cite{wu2022incorporating} & argument generation & P(Y|D,A) & 4,764&\usym{1F5F8}& \usym{1F5F4} & \usym{1F5F4} \\
ArgEssay~\cite{bao2022aeg} & argument generation & P(Y|T) & 11,282&\usym{1F5F8}& \usym{1F5F4} & \usym{1F5F4} \\
AspectCorpus~\cite{goloviznina2023aspect} & argument generation & P(Y|T,S,\textit{Asp}) & 418&\usym{1F5F4}& \usym{1F5F4} & \usym{1F5F4} \\
AIDebater~\cite{Aidebater23} & argument generation & P(Y|T,A)& -&\usym{1F5F4}& \usym{1F5F4} & \usym{1F5F4} \\
ArgSciChat~\cite{ruggeri2023dataset} & argument generation & P(Y|T,D)& 41&\usym{1F5F4}& \usym{1F5F4} & \usym{1F5F4} \\
DEBATUNE~\cite{Li2024CanLS} & argument generation & P(Y|S,A,T) & 7,100&\usym{1F5F4}& \usym{1F5F4} & \usym{1F5F8} \\ \hline
\textbf{SAD} & argument generation & P(Y|T,H,S,[R])& 392,822&\usym{1F5F8} & \usym{1F5F8} & \usym{1F5F8} \\
\bottomrule
\end{tabular}
}
\caption{Comparison between our dataset and other datasets. The corresponding abbreviations involved are as follows: Y--Response, A--Argument, T--Topic, KB--Knowledge Base, S--Stance, Asp--Aspect, D--Document, H--History, R--stRategy.}
\label{tab:com}
\end{table*}

Our contributions are summarized as follows:
\begin{itemize}[leftmargin=*]
    \item \textbf{Dataset:} We construct SAD, a large-scale, high-quality dataset for strategy-aware, multi-turn argumentation dialogue. To the best of our knowledge, SAD is the first strategy-based multi-turn argumentation dataset at this scale.
    \item \textbf{Task Formulation:} We propose a new argument generation task that better reflects real-world scenarios, requiring models to generate response arguments conditioned on the dialogue history, topics, stances, and specified strategies. SAD also supports related tasks such as general multi-turn argument generation, strategy identification, and argument retrieval.
    \item \textbf{Automatic Evaluation:} We train an automatic evaluator for argument persuasiveness and show that its judgments achieve high agreement with human ratings.
    \item \textbf{LLM Benchmarking:} We benchmark multiple LLMs and find substantial room for improvement in multi-turn argumentation. Experimental results show that incorporating argument strategies can significantly enhance training quality and generation performance.
\end{itemize}


\section{Related Work}

The Argument Generation Task is to design systems to automatically generate a persuasive text to support or refute a specific claim, 
opinion, or position. Argument datasets serve as a foundation for the design and 
evaluation of argument systems. There have been several datasets in the field of argument generation using empirical methods that focus
on generating arguments based on topics. As shown in Table~\ref{tab:com}, Gigaword~\citep{sato2015end} performs a secondary annotation
on a large news text corpus to make the dataset suitable for the argument retrieval task.
Iargument~\citep{wang2016neural} is a an argumentation dataset from \textit{iargument.org}, 
which is a Wikipedia-style website for gathering pro and con arguments on controversial issues. 
MedicalRecords~\citep{green2017argumentation} is from medical records about the patient and the 
patient’s biological family, and connections, a list of facts or principles of genetics. 
Arg-Microtexts~\citep{wachsmuth2018argumentation} is designed to provide crisp argumentation 
in a “pro and con” manner. The CMV dataset~\cite{hua2018neural}, collected from a Reddit 
subcommunity, consists of 224,553 examples with 305,475 relatively high-quality replies.
The CE2.3k dataset~\cite{gretz2020workweek} contains 1,489 examples with 2.3k manually curated 
claims extracted from Wikipedia. Each topic of CE2.3k is mapped to a corresponding Wikipedia title.
The LN55k dataset~\cite{gretz2020workweek}, collected from a corpus of newspaper articles, 
contains 224,553 examples, including 55,024 manually curated claims.
Kialo~\cite{al2021employing} contains 82,728 discussion from kialo.com, which is a argument portal.
Args.me~\cite{ajjour:2019a} includes 30,748 arguments from \textit{debatewise.org}, \textit{debate.org}, 
\textit{debatepedia.org} and \textit{idebate.org}. ~\citet{alshomary2021argument} extends 
CMV~\cite{jo2020detecting} by further collecting the quoting sentences from the 
comments (i.e., the counter-arguments). PRRCA~\cite{wu2022incorporating} is a peer review and 
rebuttal counter-arguments by collecting publicly available peer review contents and the 
submission information from \textit{openreview.net}. ArgEssay~\cite{bao2022aeg} is a
large-scale argumentative essay generation dataset.
AspectCorpus~\cite{goloviznina2023aspect} is the first russian-language corpus of arguments
with annotated aspects. AIDebater~\cite{Aidebater23} contains counter-argument generation
corpus from the ChangeMyView forum and topic argument generation from argument competition. 
ArgSciChat~\cite{ruggeri2023dataset} is a document conversation QA corpus based on scientific 
papers. DEBATUNE~\cite{Li2024CanLS} synthesizes a multi-turn argumentative dialogue dataset, 
filling the gap in multi-turn argument benchmarks. These benchmarks have greatly promoted the 
development of argument generation tasks.

Most existing argumentation corpora do not focus on the multi-turn conversational ability of 
argumentation or the controllability of generation. Although DEBATUNE~\cite{Li2024CanLS} 
is a multi-turn argumentative dialogue dataset, it is a synthetic non-real-world corpus. 
To fill this gap, this paper constructs a large-scale strategy-based multi-turn
dialogue argument dataset SAD.
\section{Dataset Creation}
This section deals with the process followed for the creation of the dataset for multi-turn
argumentative dialogue dataset. We have broadly split this into three parts:
dialogue collection, stance annotation, strategy annotation and data filtering.
\subsection{Procedure for Argument Collection}
We draw our data from the Reddit community r/ChangeMyView (CMV), which is dedicated 
to fostering open discussions on a wide range of contested issues. Specifically, CMV is structured as discussion threads, where the original post starts with a viewpoint on a controversial topic, followed by detailed supporting arguments,
then other users reply with counter-arguments. However, not all participating users counter topics they are interested in; many also express support.
During the debate, users are allowed to reply to all posts. Therefore, a topic could have multiple threads of discussion.
As shown in Figure~\ref{intro:draw_exam} (Appendix), multiple discussion threads are identified under 
the topic \textit{In a clinical setting, only medical doctors should be called "doctor"}.
Each thread represents a multi-turn debate dialogue.
\begin{table}[ht]
\begin{adjustbox}{max width=0.9\columnwidth, center}
\includegraphics[width=\columnwidth]{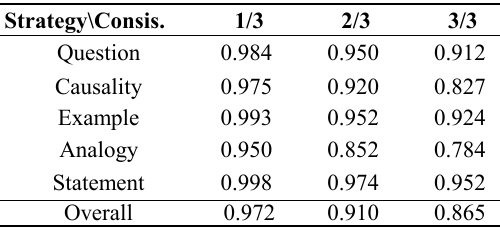}
\end{adjustbox}
\caption{Consistency proportion of strategy annotation. 1/3 means consistency with at least one 
annotator, 2/3 means consistency with at least two annotators, and 3/3 means consistency with all three annotators.}
\label{tab:cons}
\end{table}

High-quality arguments make CMV possible for constructing multi-turn argumentative dialogue 
datasets. We crawl all discussion trees created at any time from March 2016 to September 2020.
We do not account for data omissions resulting from interrupted crawling. The crawled data contains 26,726 discussion trees,
and 17,652,244 nodes. Based on the discussion threads related to each topic, we construct a 
multi-turn debate dialogue dataset.
To maintain data quality, we performed data filtering. First, we apply 
rule-based filtering to remove web links, meaningless repeated stop words, and other 
non-informative content. Posts with short utterances often fail to provide valuable 
arguments to support their points, resulting in low-quality examples. Statistical analysis 
shows that these posts typically contain fewer than 10 words per utterance. Therefore, posts 
shorter than 10 words are not considered. Specifically, given a multi-turn argument example $D$=$\{u_1,u_2,\dots,u_i,\dots,u_n\}$,
if the sentence length of $u_i$ is less than 10, then $\{u_i,u_{i+1},\dots\}$ will be removed 
from the example, and finally $D$ represents $\{u_1,u_2,\dots,u_{i-1}\}$. 
This dataset is available at \url{https://anonymous.4open.science/r/ACL_SAD}.
\subsection{Stance Annotation}
We observe that in open-ended dialogue debates, participants’ stances do not strictly 
alternate, which allows them to engage more freely and spontaneously, resulting in greater 
interactivity. This irregular progression of stances creates a more natural interactivity
dynamic, but it also causes automatic annotation to fail. We then present the process of 
annotating the stances. Specifically, we recruit and train five workers to annotate the 
answers following our guideline\footnote{All annotators in this work are compensated for 80 in
CNY per hour, which is reasonable given the difficulty of the annotation task and the mean 
income of urban residents in China.}. In this annotation task, workers are given 
topic–utterance pairs and asked to label the relationship between them, where 1 indicates 
support and 0 indicates opposition. The label with the highest number of votes is selected as 
the final answer. The Fleiss’ Kappa score for the five annotators is 0.78, indicating a high 
level of consistency and reliability in annotations.

\begin{table}[t]
    \centering
    \small
    \begin{tabular}{lr}
        \toprule
        Total Utterances & 722812.00 \\
        Total Examples & 392,822.00 \\
        Topics & 20,619 \\
        \midrule
        Avg. Utterance Length (tokens) & 119.79 \\
        Min. Utterance Length (tokens) & 11.00 \\
        Max. Utterance Length (tokens) & 2399.00 \\
        \midrule
        Max. Turns & 11.00 \\
        Min. Turns & 2.00 \\
        Avg. Turns & 3.69 \\
        Max. Dialog Length (tokens) & 9,784 \\
        Min. Dialog Length (tokens) & 24 \\
        Avg. Dialog Length (tokens) & 609.97 \\
        \bottomrule
    \end{tabular}
    \caption{Statistical information of the argument dataset}
    \label{tab:static}
\end{table}
\subsection{Strategy Annotation}
Argumentation strategies play a crucial role in determining the persuasiveness of an argument, 
when used effectively, they can dramatically enhance its overall impact. To facilitate further 
research on strategies in argument support, we then present the process that we annotate the 
argumentation strategies. 
Argumentation strategies have been systematically studied~\cite{freeley2013argumentation,weston2018rulebook}.
As shown in Table~\ref{tab:strategy},  we identify 5 strategies: Question, 
Causality, Example, Analogy and Statement according to ~\cite{freeley2013argumentation,weston2018rulebook}.
We observe that an argument may employ multiple argumentation strategies. To ensure the quality of  
the annotations, we annotate each argument and each strategy separately. Specifically, we 
recruit and train five workers with debating experience to annotate the argumentation strategies.
Each annotator's hourly compensation is the same as the previous stance annotation. In this task,
the workers are shown an argument and ask to determine whether the argument employs a particular strategy.
The workers are allowed to ignore the arguments that do not match the definition of any strategy, 
which would be automatically labeled as "None".

\subsection{Annotation Quality Control}
The stance annotation is relatively simple, and we provide Fleiss’ Kappa 
to demonstrate the reliability of the annotation. For strategy annotation, we implement 
more stringent quality control than stance annotation.
Before annotating, workers were required to go through the guideline and the provided examples.
To ensure the quality of annotations, we require workers to annotate 200 examples before the 
formal annotation, which are revised by debate professionals for feedback. We repeat the above
process until the workers are able to annotate the cases almost correctly. After annotation, to check
the quality of labels, we randomly sample 200 arguments, give them to three examiners to pick out 
incorrect labels, and calculate the consistency proportion. Results are shown in 
Table~\ref{tab:cons}. More than 97.2\% of the strategy labels are consistent with at least one examiner, 
and more than 91.0\% of the strategy labels are consistent with at least two examiners,
indicating the reliability of strategy annotation.
\section{Corpus Analysis}
The statistical information of the argument dataset is given in Table~\ref{tab:static}.
The dataset contains 392,822 dialogue instances and a total of 722,812 utterances, covering 
20,619 distinct topics, which indicates substantial scale and topical diversity. The average 
utterance length is 119.79 tokens, with a minimum of 8 tokens and a maximum of 2,399 tokens, 
reflecting a broad range of argumentative expressions from brief statements to highly detailed 
arguments. In terms of dialogue structure, each dialogue consists of 2 to 11 turns, with an 
average of 3.69 turns. Correspondingly, dialogue lengths vary widely, ranging from 24 to 9,784 
tokens, with an average length of 609.97 tokens. These statistics demonstrate that the dataset 
captures multi-turn argumentative dialogues of varying length and complexity, making it 
well-suited for training and evaluating models for argumentation understanding and generation.
\begin{figure}[htbp]
    \centering
    \includegraphics[width=0.85\columnwidth]{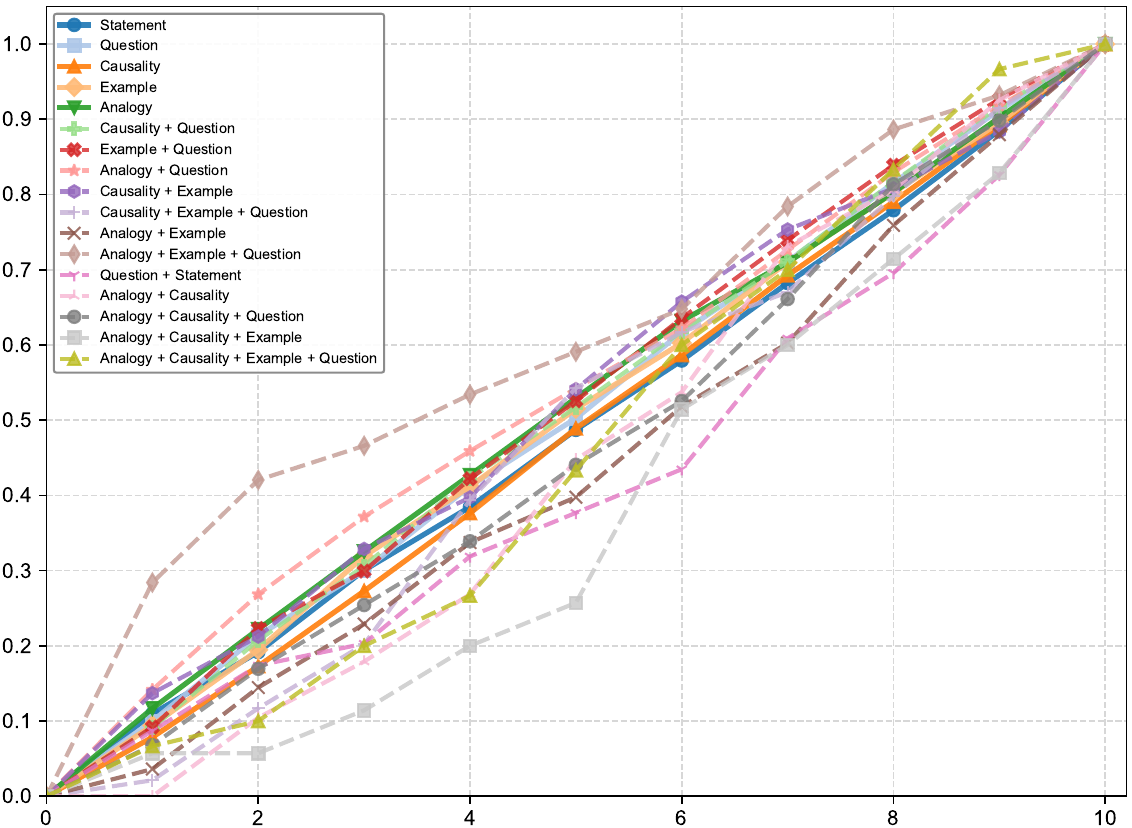}
    \caption{Cumulative distribution of strategies. The x-axis denotes the relative position in 
    dialogue, and the y-axis denotes the cumulative proportion.}
    \label{fig:cumulative1}
\end{figure}
\subsection{Strategy Sequence Analysis}
\paragraph{Cumulative Distribution of Strategies}
Figure~\ref{fig:cumulative1} presents the cumulative distribution of strategy occurrences 
across relative dialogue positions, revealing a pronounced discrepancy among different strategies.
To facilitate a clearer analysis of strategy distributions, we partition the dialogue content 
into three equal stages—beginning, middle, and ending—since strategy usage demonstrates 
distinct characteristics across different phases of the dialogue, for we observe from 
our data that most strategies have different functions and characteristics among the beginning, 
middle and ending part. For instance, \textbf{Statement} and \textbf{Question} serve as the 
dominant strategies in the early stage, showing that the debate primarily concentrates 
on establishing the argumentative focus and introducing initial claims. 

As the dialogue progresses into the middle stage, the distribution of \textbf{Causality} and 
\textbf{Example} shifts toward a steady linear growth. This phase represents the 
"logical engine" of the debate. The stabilization of these slopes reflects a consistent 
commitment to evidentiary support; participants move beyond mere assertions to establish 
mechanistic links (\textbf{Causality}) and empirical grounding (\textbf{Example}). The high 
density of overlapping curves in this region suggests a strategic equilibrium where diverse 
argumentative modes are deployed in parallel to sustain the dialectical momentum. 
At the end-stage, the strategy is not to introduce new topics but to leverage integrative 
mechanisms that weave scattered logical threads into a comprehensive, summarizing stance, 
thereby maximizing persuasiveness before the dialogue concludes. This stage typically employs 
a combination of strategies, such as the \textbf{Analogy} + \textbf{Causality} + \textbf{Example} + \textbf{Question}.

The analysis suggests a clear evolutionary trajectory of strategic complexity. The dialogue 
transitions from structural simplicity (aimed at orientation) to logical robustness 
(aimed at substantiation), and finally to syntactic complexity (aimed at integration). 
This distribution confirms that participants do not deploy strategies randomly. Rather, 
they adhere to a latent "argumentative pulse" that prioritizes foundational clarity at the 
outset and rhetorical sophistication at the conclusion.

\begin{figure}[htbp]
    \centering
    \includegraphics[width=0.85\columnwidth]{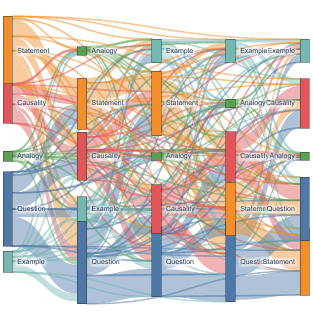}
    \caption{Visualization of the strategy flow patterns.}
    \label{fig:sankey}
\end{figure}
\subsection{Strategy Transition}
To provide deeper insights into strategy utilization, we conduct a comprehensive visualization 
analysis of strategy transitions across the first five conversational turns. As illustrated in 
Figure~\ref{fig:sankey}, the results reveal several noteworthy patterns, demonstrating that 
strategy transitions are far from random and instead follow a highly stable yet evolutionary 
framework. Transitions from declarative sentences to declarative sentences consistently account 
for the largest proportion, indicating that expressing opinions through statements remains the 
most prevalent mode of debate.

Causality emerges as a central connective strategy, exhibiting strong transitions both from 
and to Statement and Question. This indicates that causal reasoning often functions as a 
bridge between expressing claims and soliciting justification, reinforcing the logical 
coherence of arguments. Similarly, while Example and Analogy appear less frequently as 
standalone strategies, they are commonly integrated into later turns, either remaining stable 
across turns (e.g., Example → Example) or combining with core strategies to enhance explanatory 
strength.

Statements, Causality, and Question are commonly used strategies in argumentation, 
and they play a vital role in advancing the debate. Overall, the transition structure 
demonstrates that strategy usage in argumentative dialogue follows a non-random yet evolutionary 
trajectory. Speakers tend to maintain core strategies while progressively incorporating additional 
rhetorical functions, resulting in increasingly complex and information-rich argumentative 
moves.
\begin{table}[ht]
\centering
\includegraphics[width=0.80\columnwidth]{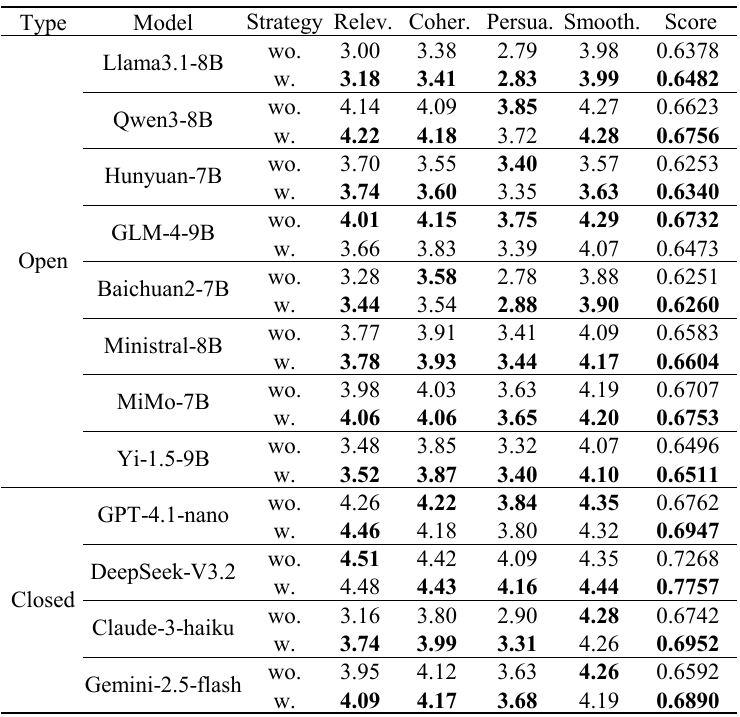}
\caption{Performance comparison of various large language models on the SAD test set using zero-shot testing. 
The Relevance (Relev.), Coherence (Coher.), Persuasiveness (Persua.), and Smoothness (Smooth.) are evaluated using GPT-4.1 as the evaluator.
All experimental results are shown in Table~\ref{tab:strategy_comparison_all}.
The score is the persuasiveness rating given by the evaluator we trained. "w." indicates the use of argumentation strategies, while "wo." indicates the absence of strategies. The corresponding prompt templates are shown in Figure~\ref{fig:prompt_for_responder_w} and Figure~\ref{fig:prompt_for_responder_wo}.}

\label{tab:strategy_comparison}
\end{table}
\section{Experiments}
The SAD dataset is divided into a training set, a validation set, and a test set in an 8:1:1 ratio. 
The experimental baselines, metrics, and settings are detailed in Appendix~\ref{app:baselines}, ~\ref{app:metrics}, and ~\ref{app:setting}.
In addition to using GPT-4.1 as the evaluator, we also train a separate persuasion evaluator, as detailed in the Appendix~\ref{app:evaluator}. The GPT-4.1 prompt template (designed in accordance with previous large language model evaluation works~\cite{liu2023g}) is provided in Figures~\ref{fig:prompt_for_gpt4_scorer_rel} to ~\ref{fig:prompt_for_gpt4_scorer_smo} (Appendix~\ref{app:temp}).
The template for the persuasion evaluator we trained is shown in Figure~\ref{fig:prompt_for_qwen3} (Appendix~\ref{app:temp}).
We provide a case study to illustrate the importance of strategy in Appendix~\ref{app:case}.

\subsection{Task Definition}
Given a topic $T$, dialogue history $H=\{u_1,u_2,\dots,u_n\}$, stance $S$, and strategy $R$, 
the debate system aims to generate a corresponding argument $A$, which can be formally 
expressed as: $P(A|T,H,S,[R])$. Here, the strategy $R$ is an optional input; in its absence, 
the task degenerates into a non-strategic multi-turn dialogue and argument generation problem.

\begin{table}[ht]
\includegraphics[width=0.95\columnwidth]{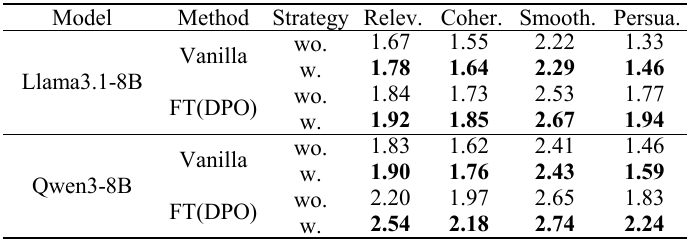}
\caption{Results of human evaluation.}
\label{tab:human}
\end{table}

\subsection{Human Evaluation}
To better evaluate the quality of the generated response and the effectiveness of the strategy,
we conduct human evaluation. We recruit five graduate students with debating experience
to conduct human evaluations. These professional raters are asked to score the generated argument in terms of 
\textbf{Smoothness}—whether the generated argument is fluent and grammatical. 
\textbf{Coherence}—whether the generated argument is logical and well
organized. 
\textbf{Relevance}—whether the descriptions in generated argument are relevant to the debate topic.
\textbf{Persuasiveness}—whether the generated argument effectively influences the audience’s beliefs or attitudes through convincing reasoning and evidence.
Each dimension is divided into three levels, corresponding to scores of 1, 2, and 3. 
For \textbf{Smoothness}: a score of 1 indicates that the generated arguments are not fluent 
and contain grammatical errors; a score of 2 indicates that the generated arguments are 
grammatically correct and generally fluent; and a score of 3 indicates that the generated 
arguments are highly fluent with no grammatical issues. The raters are asked to rate with
metrics independently.

We randomly sample 100 examples from the test set. Table~\ref{tab:human} shows the result of 
human evaluation. Overall, the results consistently demonstrate that incorporating 
argumentation strategies leads to clear performance gains across all evaluation dimensions, 
regardless of the backbone model or training regime. This trend holds for both Llama3.1-8B and 
Qwen3-8B, suggesting that the benefit of strategy awareness is largely model-agnostic.

Comparing the w. and wo. settings, we observe that strategy usage yields systematic improvements,
with the most pronounced gains in Persuasiveness and Relevance. For example, under the Vanilla 
setting, Llama3.1-8B improves from 1.67 to 1.78 in Relevance and from 1.33 to 1.46 in 
Persuasiveness when strategies are applied. Similar patterns are observed for Qwen3-8B, 
where Persuasiveness increases from 1.46 to 1.59.
These results indicate that explicit argumentation strategies help models better align their 
responses with the argumentative intent of the dialogue, making generated arguments more 
convincing and on-topic. Improvements in Coherence and Smoothness, though relatively smaller, 
further suggest that strategic planning contributes to better discourse organization.

Fine-tuning substantially enhances model performance across all metrics, and the gains from 
fine-tuning are complementary to those from strategy usage. In both models, FT + w. achieves 
the best overall results. For instance, Qwen3-8B with FT and strategies reaches 2.54 in 
Relevance and 2.24 in Persuasiveness, outperforming all other configurations.
Notably, the performance gap between w. and wo. widens after fine-tuning, especially for 
Persuasiveness. This suggests that fine-tuning enables the model to better internalize and 
operationalize argumentation strategies, rather than merely following surface-level patterns.

\subsection{Automatic Evaluation}
Table~\ref{tab:strategy_comparison} presents a multidimensional evaluation of 
open-source and closed-source large language models on the argumentation dataset, 
reporting GPT-4.1-based evaluator on Relevance, Coherence, Persuasiveness, and Smoothness, 
together with an auxiliary Score produced by an external evaluation model. 

Across both open-source and closed-source models, introducing strategy-aware inputs leads to 
consistent improvements in Relevance and Coherence. For example, Llama3.1-8B shows gains 
in Relevance (3.00 → 3.18) and Coherence (3.38 → 3.41), while Qwen3-8B improves from 4.14 to 
4.22 in Relevance and from 4.09 to 4.18 in Coherence. Similar patterns are observed across most 
models, suggesting that explicit strategy information helps models better align their responses 
with the argumentative intent and maintain logical consistency. Improvements in Smoothness are also widespread, though generally smaller in magnitude. For 
instance, MiMo-7B increases from 4.19 to 4.20, and DeepSeek-V3.2 improves from 4.35 to 4.44. 
These gains indicate that strategy guidance contributes to more structured and fluent discourse, 
even when baseline fluency is already strong.
Open-source models generally achieve moderate absolute scores across all dimensions, 
but benefit consistently from strategy-aware inputs, particularly in Relevance and Coherence. 
This indicates that the strategy provides actionable signals even for models with limited 
capacity.

At the metric level, Smoothness shows the most stable improvements, implying that strategy 
guidance contributes to better-organized and more natural argumentative responses. In contrast, 
Persuasiveness (including Persua. and Score) exhibits more heterogeneous behavior. 
While several models show noticeable improvements—such as Llama3.1-8B (2.79 → 2.83) and 
Claude-3-haiku (2.90 → 3.31)—others display marginal gains or slight decreases 
(e.g., Qwen3-8B: 3.85 → 3.72). 
This suggests that persuasive strength may require a combination of strategic awareness and 
advanced language generation abilities.
Incorporating explicit strategies leads to only limited 
gains in persuasiveness for open-source models, suggesting a restricted ability to effectively 
apply strategies. In contrast, closed-source models show substantially larger improvements in 
persuasiveness after strategy injection, indicating a stronger capability to leverage 
argumentative strategies. This phenomenon reveals meaningful differences in how models 
internalize and operationalize argumentative strategies.

Overall, the results demonstrate that the proposed dataset effectively supports fine-grained 
evaluation of argumentative quality across multiple dimensions. By revealing consistent gains 
in relevance, coherence, and fluency, while exposing the challenges of improving persuasiveness, 
the dataset provides a valuable benchmark for future research on strategy-aware argument 
generation, evaluation, and interactive argumentation dialogue systems.

\paragraph{Fine-tuning Model.}
Table~\ref{tab:ft_comparison} shows the performance of two models, Llama3.1-8B and Qwen3-8B, 
under two fine-tuning strategies: SFT (Supervised Fine-Tuning) and DPO 
(Direct Preference Optimization). In general, DPO outperforms SFT across all evaluation 
metrics—Relevance, Coherence, Persuasiveness, Smoothness, and the Score (Persuasion).
Qwen3-8B performs slightly better than Llama3.1-8B, especially under DPO, 
where both models showed significant improvements compared to their SFT versions. 
This suggests that DPO is a more effective fine-tuning strategy for tasks requiring persuasive 
and coherent responses. Therefore, for applications like debate or persuasive writing, 
DPO offers clear advantages over SFT.

The results show that using debate strategies (denoted as w.) significantly improves model 
performance compared to when these strategies are not used (wo.). Models trained with debate 
strategies perform better across most metrics—Relevance, Coherence, Persuasiveness, and 
Smoothness.

When debate strategies are applied, the models became more persuasive, with stronger arguments 
that are not only more fluent but also better structured and more relevant to the task. 
The Persuasiveness score, in particular, improved, showing that the injected debate-specific 
training helped the models create more convincing arguments.

\begin{table}[ht]
\includegraphics[width=0.95\columnwidth]{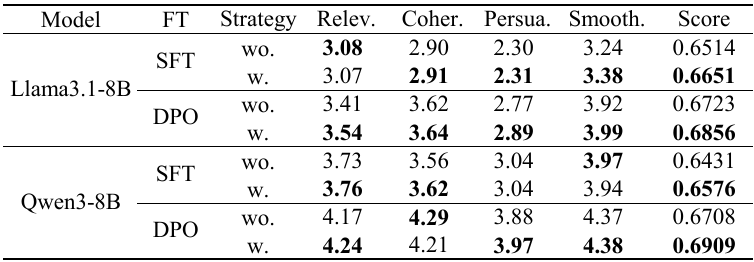}
\caption{Performance comparison across models.}
\label{tab:ft_comparison}
\end{table}

\section{Conclusion}

We introduce SAD, the first large-scale dataset for strategic multi-turn argumentation dialogues, derived from real-world interactions. 
Extensive experiments across a range of LLMs show that explicitly prompting strategies consistently improves fluency, stylistic coherence and persuasiveness.
With task-specific training, strategy cues substantially enhance the debating abilities of Open source large language model. 
We hope SAD will serve as a foundational resource for building more persuasive, strategic, and controllable dialogue systems.

\section*{Limitations}

This work has the following limitations.
Although we provide a prompt template for evaluating large language models, the evaluation of large language models is sensitive to prompts; therefore, even minor changes in the prompt may slightly perturb the evaluation results.
The dataset comes from a web forum, which inevitably introduces some biases. Models trained on this dataset may inherit these biases.

\section*{Ethical Considerations}
This paper introduces a new dataset that has been carefully processed to address ethical considerations relevant to its collection and release.
\subsection*{Privacy and Anonymization.}
All data in the dataset have undergone a thorough anonymization process. Personally identifiable information (PII), including names, user identifiers, and location-specific details, has been removed. As a result, the dataset does not contain personal or sensitive information and cannot be used to identify individuals.

\subsection*{Data Release and Intended Use}

The dataset will be publicly released to support research in the natural language processing community. It is intended for academic and non-commercial research purposes only. We encourage responsible use of the dataset and discourage applications that could cause harm, misrepresentation, or unfair treatment of individuals or social groups.

\subsection*{Potential Risks}
Although the dataset has been anonymized, it may still reflect linguistic patterns, opinions, or social biases present in the original data sources. Models trained on this dataset could inherit such biases. Users should therefore exercise caution when applying the dataset or derived models in real-world or high-stakes settings.

\bibliography{custom}

\appendix

\section{Baselines}
\label{app:baselines}
We conduct extensive testing, including \textbf{26} open-source models and \textbf{4} closed-source models.
The open-source models mainly include the \textbf{Llama3 series}~\cite{dubey2024llama} (\textit{Llama3.2-1B}, \textit{Llama3.2-3B} and \textit{Llama3.1-8B}),
\textbf{Qwen3 series}~\cite{yang2025qwen3} (\textit{Qwen3-0.6B}, \textit{Qwen3-1.7B}, \textit{Qwen3-4B}, \textit{Qwen3-8B}, \textit{Qwen3-14B} and \textit{Qwen3-32B}), \textbf{GLM series}~\cite{glm2024chatglm} (\textit{GLM4-9B} and \textit{GLM4-32B}), 
\textbf{Gemma series}~\cite{gemma_2025} (\textit{Gemma3-1B}, \textit{Gemma3-4B}, \textit{Gemma3-12B} and \textit{Gemma3-27B}), \textbf{Baichuan series}~\cite{baichuan2023baichuan2} (\textit{Baichuan2-7B} and \textit{Baichuan2-13B}), 
\textbf{Yi series}~\cite{young2024yi} (\textit{Yi1.5-6B} and \textit{Yi1.5-9B}), \textbf{Hunyuan series}~\cite{hunyuan2025} (\textit{Hunyuan-4B} and \textit{Hunyuan-7B}), \textbf{Phi series}~\cite{abdin2024phi} (\textit{Phi-4} and \textit{Phi-4-mini}), 
\textbf{Ministral-8B}~\citep{mistralai2024ministral}, \textbf{MiMo-7B}~\cite{xiaomi2025mimo}, and \textbf{GPT-OSS-20b}~\cite{agarwal2025gpt}. The closed-source models mainly include \textbf{GPT-4.1}(nano)~\cite{hurst2024gpt}, \textbf{DeepSeek}-V3.2~\cite{liu2024deepseek}, \textbf{Claude-3}(haiku)~\cite{anthropic2024claude3} and \textbf{Gemini}-2.5(flash)~\cite{comanici2025gemini}.

\section{Metrics}
\label{app:metrics}
Traditional reference-based evaluation 
metrics, such as BLEU and ROUGE, cannot accurately reflect the quality of the generated text~\cite{yao2024clave,liu2023evaluate,hu2024llm}.
Compared to reference-based evaluation metrics, reference-free metrics have been shown to have 
better properties related to human preferences~\cite{liu2023g,yao2024clave}. First, we use a 
GPT-4.1-based evaluator to assess the quality of the generated text in terms of \textbf{Relevance}, 
\textbf{Coherence}, \textbf{Smoothness}, and \textbf{Persuasiveness}. 
\textbf{Relevance} refers to the extent to which a response directly addresses the topic and 
meaningfully engages with the debate history. The response should stay on topic, acknowledge 
and respond to previous arguments, and contribute relevant points to theongoing discussion.
\textbf{Coherence} refers to the extent to which a response be well-organized, with clear 
connections between ideas. It should build from point to point in a logical manner, 
not just present a collection ofdisconnected statements.
\textbf{Smoothness} refers to the extent to which a response be naturally written, easy to 
follow, and free from awkward phrasing or grammatical issues. The language should flow smoothly 
and maintainappropriate tone for a debate context.
\textbf{Persuasiveness} refers to the extent to which a response presents compelling arguments, 
provides adequate reasoning or evidence to support claims, and effectively advance its position 
in the debate.

To effectively evaluate the persuasiveness of generated arguments, we train a persuasiveness 
estimator based on the number of likes received by users. Please refer to Appendix~\ref{app:evaluator} for detailed 
procedures.

\section{Persuasiveness Evaluator}
\label{app:evaluator}
Drawing on the collected corpus, we propose a multi-agent framework for evaluating the
persuasiveness of argumentation systems. Specifically, given a topic $T$, a stance $S$,
a dialogue history $H$, an optional argumentation strategy $R$, and a generated argument $A$,
the evaluator is tasked with scoring the persuasiveness of $A$ conditioned on $T$, $H$, $S$,
and $[R]$. This evaluation process is formalized as: $P(\text{Score} \mid H, T, S, [R], A)$
The score corresponds to the number of likes an argument receives from users. Prior to
training, we normalize the scores to the range $[0,1]$. Using the processed data, we train
the evaluators via supervised fine-tuning.

To assess the reliability of the proposed evaluator, we measure its consistency with human
judgments, using the actual voting scores from the CMV platform as ground truth. As shown
in Table~\ref{tab:evaluator_performance}, the Qwen3-8B-based evaluator demonstrates strong
alignment with human evaluations, achieving a Pearson correlation coefficient of $0.71$
and a Spearman correlation coefficient of $0.66$. Accordingly, we adopt the Qwen3-8B-based
evaluator as a reference metric for assessing the persuasiveness of generated arguments.

\begin{table}[htbp]
\centering
\begin{tabular}{lcc}
\toprule
\textbf{Model} & \textbf{r} & \textbf{$\boldsymbol{\rho}$}\\
\midrule
Qwen3-0.6B   & 0.64 & 0.60   \\
Llama3.2-1B  & 0.56 & 0.53   \\
Qwen3-1.7B   & 0.67 & 0.62  \\
Qwen3-4B     & 0.70 & 0.64   \\
Qwen3-8B     & 0.71 & 0.66  \\
\bottomrule
\end{tabular}
\caption{Spearman($\rho$) and Pearson(r) correlation between different evaluators and human 
like scores. }
\label{tab:evaluator_performance}
\end{table}

\section{Experiments Details}
\label{app:setting}

All experiments are conducted on four NVIDIA A100 GPUs. For the training phase, we utilized the LLaMA Factory framework. For training the evaluator, 
we employed Low-Rank Adaptation (LoRA) fine-tuning for the 8B model, while full-parameter 
fine-tuning was applied to all other models. For training the dialogue systems, LoRA fine-tuning 
was uniformly adopted across all model sizes. The LoRA configuration was kept consistent 
throughout, with rank set to 128 and alpha set to 256.
All inference processes are performed using vLLM. 

\begin{table}[htbp]
\centering
\small
\begin{tabular}{l|cc}
\toprule
\textbf{Parameter} & \textbf{Debater} & \textbf{Evaluator} \\
\midrule
Hardware & \multicolumn{2}{c}{4 $\times$ NVIDIA A100} \\
\midrule

\multicolumn{3}{c}{\textbf{Training (Llama-Factory)}} \\
\midrule
Training Stage & DPO & SFT \\
Fine-tuning Method & LoRA & Full \\
Learning Rate & 1.0e-5 & 1.0e-5 \\
Epochs & 3 & 3 \\
Max Model Length & 32,768 & 32,768 \\
Batch Size & 4 & 16 \\
Gradient Accumulation & 4 & 16 \\

\multicolumn{3}{l}{\textit{\quad $\hookrightarrow$ \textbf{LoRA (Debater)}}} \\
\quad Rank ($r$) & 128 & N/A \\
\quad Alpha ($\alpha$) & 256 & N/A \\
\quad Target Modules & All & N/A \\

\multicolumn{3}{l}{\textit{\quad $\hookrightarrow$ \textbf{DPO (Debater)}}} \\
\quad Beta ($\beta$) & 0.1 & - \\
\quad Loss Type & Sigmoid & - \\

\midrule

\multicolumn{3}{c}{\textbf{Inference (vLLM)}} \\
\midrule
Temperature & 0.95 & 0 (Greedy) \\
Top-$p$ & 0.7 & 1.0 \\
Top-$k$ & 50 & N/A \\
Repetition Penalty & 1.0 & 1.0 \\
Max New Tokens & 8,240 & 2048 \\
\bottomrule
\end{tabular}
\caption{Detailed hyperparameters for training and inference phases using Llama-Factory and vLLM.}
\label{tab:hyperparameters}
\end{table}

\paragraph{Preference Dataset Construction} To enhance the model's ability to discriminate between persuasive and mediocre arguments, we construct a preference dataset $\mathcal{D}_{\text{DPO}}$ to facilitate Direct Preference Optimization.

We traverse the debate tree for each topic to identify branching nodes where a single dialogue history $H$ elicits multiple sibling responses $R = \{r_1, r_2, \dots, r_n\}$ (where $n \ge 2$). We identify the branching node corresponding to the pair of responses that exhibits the largest difference in community votes. The response with the higher score is designated as chosen ($A_w$), and the one with the lower score as rejected ($A_l$).

This ensures that the DPO training is driven by the strongest preference signals, allowing the model to learn the subtle features that distinguish highly persuasive arguments from less effective ones.

\paragraph{Annotator Population}
We report the basic demographic and geographic characteristics of the annotator population. Specifically, most of the annotators are from Asia, with 60\% identifying as male and 40\% as female, and an age range of 25-36 years. Geographically, the annotators were primarily from East Asia, which reflects the target population for our study. These details are included to provide transparency regarding the composition of the annotator pool and to help interpret the annotation outcomes in the context of potential demographic or geographic biases.
\section{Templates}

\paragraph{Instruction for Evaluator} 

To assess the persuasiveness of arguments using our self-trained evaluator, we designed a structured prompt that facilitates batch processing of dialogue paths. As illustrated in Figure~\ref{fig:prompt_for_qwen3}, the input consists of the \textit{Topic}, the \textit{Debate History}, and the \textit{Responses} to be evaluated. Our evaluator is instructed to assign a direct, continuous score ranging from 0 to 1 for each response, where 0 represents poor argumentation and 1 represents highly persuasive argumentation. To ensure robust parsing of the model's output, we enforce a strict XML-style format, requiring the model to enclose the numerical score within \texttt{<answer>} tags. This design allows the evaluator to score all branching responses of a complete dialogue path simultaneously, improving evaluation efficiency.

\paragraph{Instruction for GPT-4.1 Evaluator}
To ensure a comprehensive and scalable assessment of the model's performance in multi-turn debate scenarios, we adopt the "LLM-as-a-Judge" paradigm. Specifically, we employ GPT-4.1 as the automated evaluator.

We designed four instruction prompts to evaluate the generated responses across four distinct dimensions: Relevance, Coherence, Persuasiveness, Smoothness. For each dimension, the evaluator is provided with the Topic, the Debate History and the model's Response, and is tasked with assigning a score on a scale from 1 to 5. The detailed prompts uesed for these evaluations are presented in Figure~\ref{fig:prompt_for_gpt4_scorer_rel}, Figure~\ref{fig:prompt_for_gpt4_scorer_coh}, Figure~\ref{fig:prompt_for_gpt4_scorer_per} and Figure~\ref{fig:prompt_for_gpt4_scorer_smo}.

\paragraph{Instruction for Dialogue System} 

We employ two distinct prompt configurations to investigate the impact of explicit strategy guidance on argument generation. As shown in Figure~\ref{fig:prompt_for_responder_w}, the template of generation with strategy incorporates a specific \textit{Strategy} field alongside the \textit{Topic} and \textit{Debate History}. The instruction explicitly constrains the model to adopt the specified argumentative strategy. Figure~\ref{fig:prompt_for_responder_wo} shows the template of generation without strategy, providing identical context (Topic and History) but omits the strategy constraint.

In both cases, we adopt a concise prompting approach that prioritizes the dialogue context, ensuring the model generates a definite and contextually aligned response.

\label{app:temp}
\begin{figure*}[t]
\begin{tcolorbox}[
    colframe = gray,       
    colback = gray!5!white,             
    coltitle = white,                   
    coltext = black,                   
    fonttitle = \bfseries,              
    title = {Instruction for Evaluator (Qwen3-8B)},  
    boxrule = 0.5pt,                    
    arc = 2mm,                        
    width = \textwidth,               
    left = 7pt,                        
    right = 7pt,                        
    top = 5pt,                          
    bottom = 5pt                        
]
\fontsize{10pt}{10pt}\selectfont
You are an expert in the field of argumentation. Now, for a given \# Topic and corresponding 
\# Debate History, you give a direct score for the response based on persuasiveness, ranging from 0 to 1, 
indicating very poor argumentation to very good argumentation. Your responses will be used for research purposes only, 
so please have a definite answer.\\[\baselineskip]

You should format the response using the following structure, with [SEP] between each rating:

\texttt{<answer>0.5</answer>[SEP]<answer>1.0</answer>} (only a direct number between 0 and 1)\\[\baselineskip]

\textbf{\# Topic}\\
\{topic\}\\[\baselineskip]

\textbf{\# Debate history}\\
\{context\_1\}:{sentence1}

\{context\_2\}:{sentence2}

\textbf{\# Response}\\
\{context\}:{response}

……

\end{tcolorbox}
\caption{Instruction for Evaluator (Qwen3-8B). A self-trained evaluator is used for persuasiveness assessment.}
\label{fig:prompt_for_qwen3}
\end{figure*}

\begin{figure*}[t]
\begin{tcolorbox}[
    colframe = gray,       
    colback = gray!5!white,             
    coltitle = white,                   
    coltext = black,                   
    fonttitle = \bfseries,              
    title = {Instruction for GPT-4.1 Evaluator (Relevance)},  
    boxrule = 0.5pt,                    
    arc = 2mm,                        
    width = \textwidth,               
    left = 7pt,                        
    right = 7pt,                        
    top = 5pt,                          
    bottom = 5pt                        
]
\fontsize{9pt}{11pt}\selectfont
For a given \# Response, your task is to rate the Response according to \# Topic 
and corresponding \# Debate History.
Please make sure you read and understand these instructions carefully. Please keep this 
document open while reviewing, and refer to it as needed.

\textbf{Evaluation Criteria:}

\textbf{Relevance (1-5)} - whether the Response directly addresses the Topic and engages meaningfully with the Debate History. The Response should stay on Topic, acknowledge and respond to previous arguments, and contribute relevant points to the ongoing discussion.

\textbf{Evaluation Steps:}\\
1. Read the Topic and Debate History carefully to understand the context and previous arguments.\\
2. Read the response thoroughly.\\
3. Evaluate the response on the dimensions: Relevance.\\
4. Assign scores from 1 (lowest) to 5 (highest) based on the evaluation dimension.

\textbf{Topic:}\\
\{topic\}

\textbf{Debate History:}\\
\{debate\_history\}

\textbf{Response:}\\
\{model\_prediction\}

\textbf{Evaluation Form (scores ONLY):}\\
- Relevance:\\
\end{tcolorbox}
\caption{Instruction for GPT-4.1 Evaluator (Relevance).}
\label{fig:prompt_for_gpt4_scorer_rel}
\end{figure*}

\begin{figure*}[t]
\begin{tcolorbox}[
    colframe = gray,       
    colback = gray!5!white,             
    coltitle = white,                   
    coltext = black,                   
    fonttitle = \bfseries,              
    title = {Instruction for GPT-4.1 Evaluator (Coherence)},  
    boxrule = 0.5pt,                    
    arc = 2mm,                        
    width = \textwidth,               
    left = 7pt,                        
    right = 7pt,                        
    top = 5pt,                          
    bottom = 5pt                        
]
\fontsize{9pt}{11pt}\selectfont
For a given \# Response, your task is to rate the Response according to \# Topic 
and corresponding \# Debate History.
Please make sure you read and understand these instructions carefully. Please keep this 
document open while reviewing, and refer to it as needed.

\textbf{Evaluation Criteria:}

\textbf{Coherence (1-5)} - the logical flow and structure of the argument. The Response should be well-organized, with clear connections between ideas. It should build from point to point in a logical manner, not just present a collection of disconnected statements.\\[\baselineskip]

\textbf{Evaluation Steps:}\\
1. Read the Topic and Debate History carefully to understand the context and previous arguments.\\
2. Read the response thoroughly.\\
3. Evaluate the response on the dimensions: Coherence.\\
4. Assign scores from 1 (lowest) to 5 (highest) based on the evaluation dimension.

\textbf{Topic:}\\
\{topic\}

\textbf{Debate History:}\\
\{debate\_history\}

\textbf{Response:}\\
\{model\_prediction\}

\textbf{Evaluation Form (scores ONLY):}\\
- Relevance:\\
\end{tcolorbox}
\caption{Instruction for GPT-4.1 Evaluator (Coherence).}
\label{fig:prompt_for_gpt4_scorer_coh}
\end{figure*}

\begin{figure*}[t]
\begin{tcolorbox}[
    colframe = gray,       
    colback = gray!5!white,             
    coltitle = white,                   
    coltext = black,                   
    fonttitle = \bfseries,              
    title = {Instruction for GPT-4.1 Evaluator (Persuasiveness)},  
    boxrule = 0.5pt,                    
    arc = 2mm,                        
    width = \textwidth,               
    left = 7pt,                        
    right = 7pt,                        
    top = 5pt,                          
    bottom = 5pt                        
]
\fontsize{9pt}{11pt}\selectfont
For a given \# Response, your task is to rate the Response according to \# Topic 
and corresponding \# Debate History.
Please make sure you read and understand these instructions carefully. Please keep this 
document open while reviewing, and refer to it as needed.

\textbf{Evaluation Criteria:}

\textbf{Persuasiveness (1-5)} - the strength and effectiveness of the argumentation. The Response should present compelling arguments, provide adequate reasoning or evidence to support claims, and effectively advance its position in the debate.\\[\baselineskip]

\textbf{Evaluation Steps:}\\
1. Read the Topic and Debate History carefully to understand the context and previous arguments.\\
2. Read the response thoroughly.\\
3. Evaluate the response on the dimensions: Persuasiveness.\\
4. Assign scores from 1 (lowest) to 5 (highest) based on the evaluation dimension.

\textbf{Topic:}\\
\{topic\}

\textbf{Debate History:}\\
\{debate\_history\}

\textbf{Response:}\\
\{model\_prediction\}

\textbf{Evaluation Form (scores ONLY):}\\
- Relevance:\\
\end{tcolorbox}
\caption{Instruction for GPT-4.1 Evaluator (Persuasiveness).}
\label{fig:prompt_for_gpt4_scorer_per}
\end{figure*}

\begin{figure*}[t]
\begin{tcolorbox}[
    colframe = gray,       
    colback = gray!5!white,             
    coltitle = white,                   
    coltext = black,                   
    fonttitle = \bfseries,              
    title = {Instruction for GPT-4.1 Evaluator (Smoothness)},  
    boxrule = 0.5pt,                    
    arc = 2mm,                        
    width = \textwidth,               
    left = 7pt,                        
    right = 7pt,                        
    top = 5pt,                          
    bottom = 5pt                        
]
\fontsize{9pt}{11pt}\selectfont
For a given \# Response, your task is to rate the Response according to \# Topic 
and corresponding \# Debate History.
Please make sure you read and understand these instructions carefully. Please keep this 
document open while reviewing, and refer to it as needed.

\textbf{Evaluation Criteria:}

\textbf{Smoothness (1-5)} - the linguistic fluency and readability of the Response. The Response should be naturally written, easy to follow, and free from awkward phrasing or grammatical issues. The language should flow smoothly and maintain appropriate tone for a debate context.\\[\baselineskip]

\textbf{Evaluation Steps:}\\
1. Read the Topic and Debate History carefully to understand the context and previous arguments.\\
2. Read the response thoroughly.\\
3. Evaluate the response on the dimensions: Smoothness.\\
4. Assign scores from 1 (lowest) to 5 (highest) based on the evaluation dimension.

\textbf{Topic:}\\
\{topic\}

\textbf{Debate History:}\\
\{debate\_history\}

\textbf{Response:}\\
\{model\_prediction\}

\textbf{Evaluation Form (scores ONLY):}\\
- Relevance:\\
\end{tcolorbox}
\caption{Instruction for GPT-4.1 Evaluator (Smoothness).}
\label{fig:prompt_for_gpt4_scorer_smo}
\end{figure*}

\begin{figure*}[t]
\begin{tcolorbox}[
    colframe = gray,       
    colback = gray!5!white,             
    coltitle = white,                   
    coltext = black,                   
    fonttitle = \bfseries,              
    title = {Instruction for Generation With Strategy},  
    boxrule = 0.5pt,                    
    arc = 2mm,                        
    width = \textwidth,               
    left = 7pt,                        
    right = 7pt,                        
    top = 5pt,                          
    bottom = 5pt                        
]
\fontsize{11pt}{8pt}\selectfont
\textbf{\# Topic}\\
\{topic\}\\[\baselineskip]

\textbf{\# Debate History}\\
\{context\_1\}:{sentence1}

\{context\_2\}:{sentence2}

……\\[\baselineskip]

You are an expert of debate, based on this \#Topic and \#Debate History, use the following 
specified \{\#Strategy\} strategy to conduct a discussion. Your responses will be used for research purposes 
only, so please have a definite reply.\\[\baselineskip]

\textbf{\# strategy}\\
\{strategy\_str\}
\end{tcolorbox}
\caption{Instruction for generation with strategy. Refer to Table~\ref{tab:strategy} for strategies and strategy descriptions.}
\label{fig:prompt_for_responder_w}
\end{figure*}

\begin{figure*}[t]
\begin{tcolorbox}[
    colframe = gray,       
    colback = gray!5!white,             
    coltitle = white,                   
    coltext = black,                   
    fonttitle = \bfseries,              
    title = {Instruction for Generation without Strategy},  
    boxrule = 0.5pt,                    
    arc = 2mm,                        
    width = \textwidth,               
    left = 7pt,                        
    right = 7pt,                        
    top = 5pt,                          
    bottom = 5pt                        
]
\fontsize{11pt}{8pt}\selectfont
\textbf{\# Topic}\\
\{topic\}\\[\baselineskip]

\textbf{\# Debate History}\\
\{context\_1\}:{sentence1}

\{context\_2\}:{sentence2}

……\\[\baselineskip]

You are an expert of debate, based on this \#Topic and \#Debate History, please conduct a discussion. Your responses will be used for research purposes only, so please have a definite reply.

\end{tcolorbox}
\caption{Instruction for Generation without Strategy}
\label{fig:prompt_for_responder_wo}
\end{figure*}

\begin{table*}[ht]
\begin{adjustbox}{max width=0.95\textwidth, center}
\includegraphics[width=\textwidth]{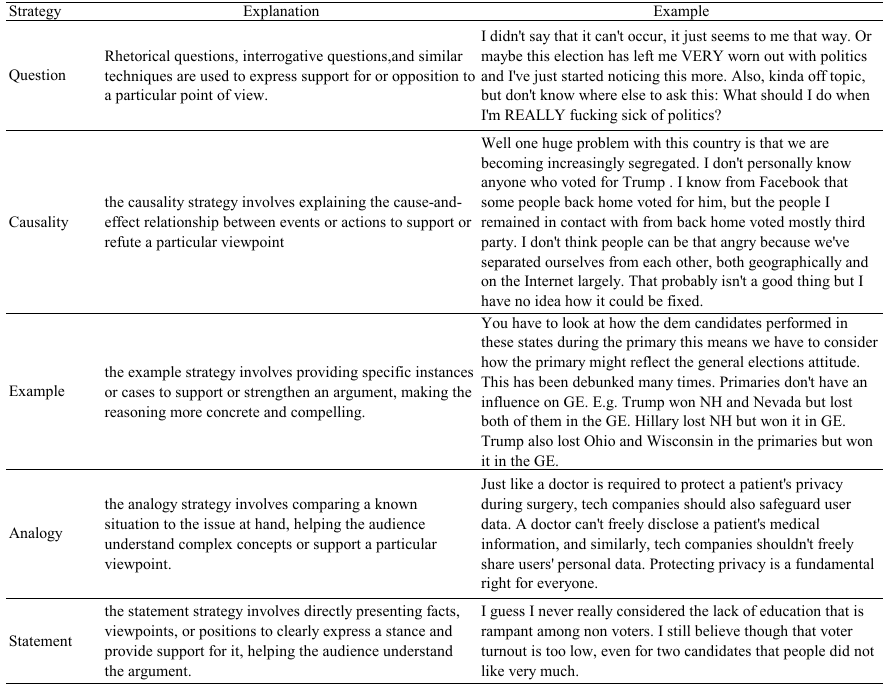}
\end{adjustbox}
\caption{Argument Strategy.}
\label{tab:strategy}
\end{table*}

\section{More Results}

Table~\ref{tab:strategy_comparison_all} provides a comprehensive evaluation of 30 Large Language Models, ranging from lightweight models (e.g., Qwen3-0.6B, Llama3.2-1B) to large-scale open-weights and closed-source models. We observe some key trends across this extensive benchmark.

Consistent with the selected results in Table~\ref{tab:strategy_comparison}, the introduction of explicit argumentation strategies (the "w." setting) yields broad performance improvements across nearly all model families and sizes, suggesting that strategic prompts effectively constrain the generation space, preventing smaller models from hallucinating or drifting off-topic.

The comprehensive data reveals a clear correlation between model scale and the ability to effectively operationalize argumentation strategies. Within the \textit{Qwen3} family (0.6B to 32B), the absolute gains in the automated \textbf{Score} metric generally widen as parameter count increases.

While closed-source models like \textit{DeepSeek-V3.2} and \textit{GPT-4.1-nano} achieve the highest overall scores (reaching up to 0.7757), top-tier open-source models demonstrate remarkable competitiveness. Specifically, \textit{Gemma-3-27B} and \textit{Qwen3-32B} achieve performance levels comparable to, and in some metrics exceeding, closed-source counterparts.

\label{app:results}

\begin{table*}[ht]
\resizebox{0.65\textwidth}{!}{\includegraphics{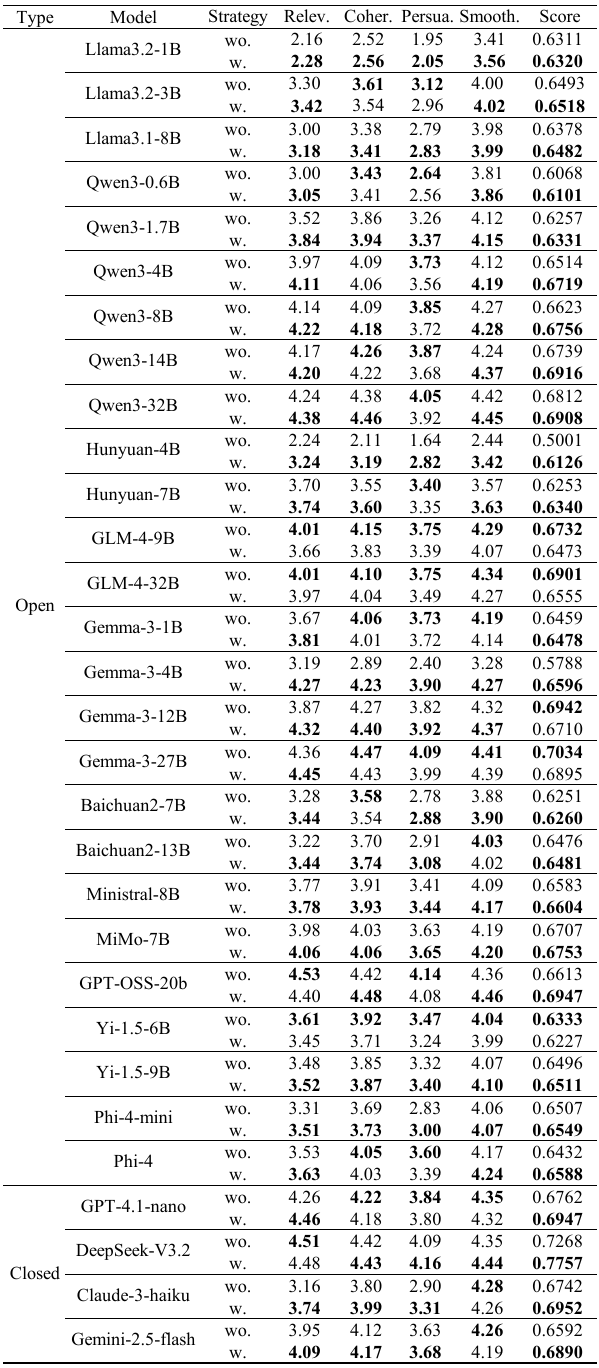}}
\centering
\caption{Performance comparison of various large language models on the SAD test set using zero-shot testing. The prompt template is shown in Figure~\ref{fig:prompt_for_responder_w} and Figure~\ref{fig:prompt_for_responder_wo}.}
\label{tab:strategy_comparison_all}
\end{table*}

\section{Case Study}
\label{app:case}
\begin{figure*}[!t]
\centering
\includegraphics[width=0.95\textwidth]{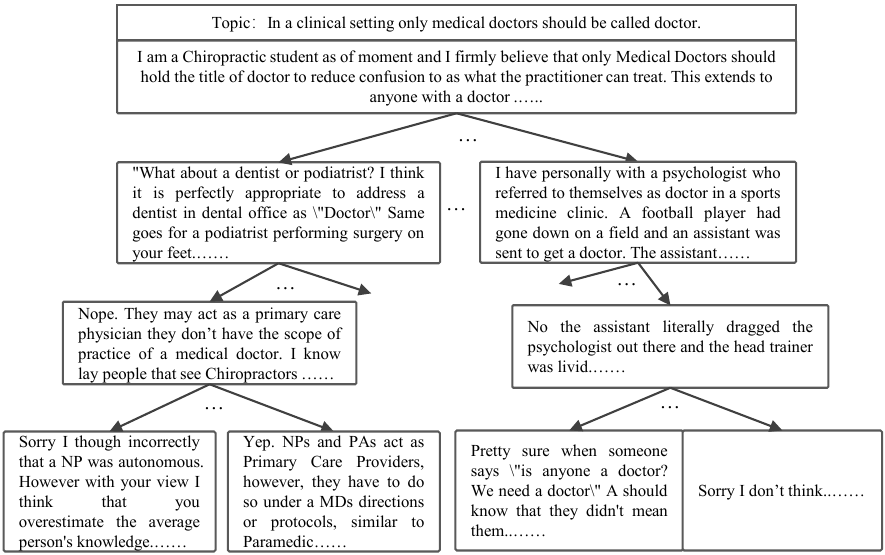}
\caption{An overview of the CMV multithreading argument.}
\label{intro:draw_exam}
\end{figure*}

As illustrated in Figure~\ref{intro:draw_exam} and Figure~\ref{intro:example}, the structure and content of a topic's data are clearly revealed. Figure~\ref{intro:draw_exam} presents the data from a topic-centric perspective, exhibiting a multi-way tree structure. From the original poster's standpoint, after presenting a topic at the outset, they subsequently provide their own viewpoint. For other participants, they engage in argumentation targeting specific responses, thereby forming distinct paths, where each path can be regarded as an independent multi-turn dialogue. Figure~\ref{intro:example} displays a concrete dialogue path, where strategic sentences are emphasized using colors other than black. For the model acting as the Responder, it needs to participate in argumentation at a particular position and develop a new branch from the existing thread. For the model serving as the Assistant, it needs to integrate the topic and the dialogue history up to the Responder's reply to assign scores to all responses along this path.

\paragraph{A Specific Comparison}
Table~\ref{tab:case_study} illustrates the qualitative impact of explicit argumentation strategies. The model utilizing the \textit{Example and Question} strategy demonstrates a more nuanced and persuasive approach. By citing specific historical precedents (e.g., the Help America Vote Act and \textit{Shelby County v. Holder}) and posing critical questions regarding the balance between federal standardization and state autonomy, the model effectively engages the user in a dialectical process. This approach not only grounds the argument in reality but also guides the user to reconsider the practicality of the proposal. In contrast, the model generated without a specific strategy produces a structured yet static response. While it covers similar informational ground (Legal Foundation, Advantages), it adopts a declarative, essay-like tone that lacks the rhetorical engagement necessary to address the user's specific doubts, highlighting the efficacy of strategic guidance in enhancing argumentative depth.

\begin{table*}[htbp]
  \begin{center}
  \begin{tabular}{p{0.95\textwidth}}
  \toprule
  \textbf{Topic:} congress has the power to impose strict regulations on federal elections, and should utilize it to standardize as well as publicize results. \\
  \textbf{1ds1m7:} My opinion is two fold. The first is that in article 1, section 4... the Congress may at any time by Law make or alter such Regulations... Congress has used this power many times. For example, the Voting Rights Act, the Help America Vote Act... The precedent exists for the Congress to legislate elections within the state. ... In the modern era, technology has both allowed us to standardize many things... I believe that elections are one such process... Congress should exercise this power. ... Federal intervention can reconcile these issues by standardizing the entire process. Furthermore, voter identification doesn't have to be tied to individual ballots... Things that would change my view would be case law that denies Congress the power to do this... \\
  \midrule 
  \textbf{Human:} \hlqu{Could you clarify what Congress should do to "standardize" elections?} \hlex{You mention voter ID laws as an example of something Congress could do,} but given how much disagreement there is... I'm not sure how easy or practical that would be.\\
  \midrule 
  \textbf{GPT4.1-nano with strategy:} I concur that Congress possesses the constitutional authority... Historical precedents such as the Voting Rights Act of 1965 and the Help America Vote Act of 2002 demonstrate Congress's capacity to intervene... \hlex{For example, the Help America Vote Act established minimum election administration standards... leading to more consistent and reliable election results nationwide.}\\
  Given the technological advancements... Standardization can reduce discrepancies... \hlex{For instance, inconsistent voter ID laws across states have created confusion... federal standards could address these disparities.} \\
  \hlqu{However, a critical question arises: how do we balance federal standardization with states' rights and local electoral autonomy?} ... Yet, does this mean that state-level variations should continue to undermine the legitimacy of federal elections? ...\\
  \hlqu{Furthermore, how can Congress ensure that standardization does not lead to overreach or disenfranchisement?} \hlex{For example, in cases like Shelby County v. Holder (2013)...} \hlbo{Would similar concerns apply to broad federal regulation today, and how can they be mitigated?}\\
  In conclusion... it must do so with careful consideration... \hlqu{Would establishing clear, constitutionally grounded guidelines that respect state autonomy while promoting uniformity be the optimal approach?} \\
  \midrule 
  \textbf{GPT4.1-nano without strategy:} Based on the topic... there are several key points to consider. \\
  **Legal Foundation and Historical Precedent:** \\
  The Constitution... grants Congress the authority... Historically, Congress has exercised this power through legislation such as the Voting Rights Act... This suggests that federal oversight is both constitutionally grounded...\\
  **Advantages of Federal Regulation:** \\
  1. **Standardization:** Federal regulations can create uniform standards... For example, standardized voter identification procedures...  \\
  2. **Public Confidence:** ...  \\
  3. **Efficiency and Technology:** ...\\
  **Addressing Common Concerns:** \\
  Some argue that federal regulation might infringe on states' rights... However, these concerns can be mitigated... For instance, federal laws can specify minimal identification requirements...\\
  **Conclusion:** \\
  Given the constitutional authority... it is both appropriate and necessary for Congress to impose strict, standardized regulations...\\
  \bottomrule
  \end{tabular}
  \end{center}
  \caption{Case Study of Argumentation Strategies. The highlighted text indicates the specific strategy used: \colorbox{c_example}{Example}, \colorbox{c_question}{Question}, and \colorbox{c_both}{Combined}.}
  \label{tab:case_study}
\end{table*}

\end{document}